\def\year{2022}\relax
\documentclass[letterpaper]{article} 
\usepackage{aaai22}  
\usepackage{times}  
\usepackage{helvet}  
\usepackage{courier}  
\usepackage[hyphens]{url}  
\usepackage{graphicx} 
\usepackage{natbib}  
\usepackage{caption} 
\DeclareCaptionStyle{ruled}{labelfont=normalfont,labelsep=colon,strut=off} 
\usepackage{algorithm}
\usepackage{algorithmic}
\usepackage{tikz}
\usepackage{booktabs}
\usepackage{xcolor,colortbl}
\usepackage{newfloat}
\usepackage{listings}
\usepackage{afterpage,tabularx}
\usepackage{mathtools}

\usepackage[shortlabels,inline]{enumitem}
\floatstyle{ruled}
\newfloat{listing}{tb}{lst}{}
\floatname{listing}{Listing}
\title{Finding Backdoors to Integer Programs: A Monte Carlo Tree Search Framework}
\author {
    Elias B. Khalil\textsuperscript{\rm 1, 2},
    Pashootan Vaezipoor\textsuperscript{\rm 3, 4},
    Bistra Dilkina\textsuperscript{\rm 5}
}
\affiliations {
    \textsuperscript{\rm 1}Department of Mechanical \& Industrial Engineering, University of Toronto\\
    \textsuperscript{\rm 2}Scale AI Research Chair in Data-Driven Algorithms for Modern Supply Chains\\
    \textsuperscript{\rm 3}Department of Computer Science, University of Toronto\\
    \textsuperscript{\rm 4}Vector Institute for Artificial Intelligence\\
    \textsuperscript{\rm 5}Department of Computer Science, University of Southern California\\
    khalil@mie.utoronto.ca, pashootan@cs.toronto.edu, dilkina@usc.edu

}

\usepackage{subfigure}

\usepackage{amsfonts}
\usepackage{amsmath}
\usepackage{amsthm}
\usepackage{mathtools}
\usepackage{amssymb}
\usepackage{xspace}
\newcommand{\bamcts}{\texttt{BaMCTS}\xspace}
\newcommand{\biased}{\texttt{Biased}}
\newcommand{\setcover}{\texttt{SetCover}}
\usepackage{xcolor}

\urldef\myurl\url{http://www.dei.unipd.it/~fisch/papers/slides/2014%20ISCO%20%5bplenary%20Fischetti%20BRANCHstorming%5d.pdf }

\newcommand{\floor}[1]{\lfloor #1 \rfloor}
\newcommand{\ceil}[1]{\lceil #1 \rceil}

\usepackage{array,multirow}
\newcommand{\STAB}[1]{\begin{tabular}{@{}c@{}}#1\end{tabular}}

\begin{document}

\maketitle

\begin{abstract}
In Mixed Integer Linear Programming (MIP), a (strong) \textit{backdoor} is a ``small" subset of an instance's integer variables with the following property: in a branch-and-bound procedure, the instance can be solved to global optimality by branching \textit{only} on the variables in the backdoor.
Constructing datasets of pre-computed backdoors for widely used MIP benchmark sets or particular problem families can enable new questions around novel structural properties of a MIP, or explain why a problem that is hard in theory can be solved efficiently in practice. Existing algorithms for finding backdoors rely on sampling candidate variable subsets in various ways, an approach which has demonstrated the existence of backdoors for some instances from MIPLIB2003 and MIPLIB2010. However, these algorithms fall short of consistently succeeding at the task due to an imbalance between exploration and exploitation. We propose \bamcts, a Monte Carlo Tree Search framework for finding backdoors to MIPs. Extensive algorithmic engineering, hybridization with traditional MIP concepts, and close integration with the CPLEX solver have enabled our method to outperform baselines on MIPLIB2017 instances, finding backdoors more frequently and more efficiently. 
\end{abstract}

\section{Introduction}
Hard discrete optimization problems arise in a very wide range of application domains. While the theoretical computer science approach focuses on the design of algorithms with approximation guarantees, many real problems are not amenable to such analyses. In the artificial intelligence and operations research communities, on the other hand, much of the focus is on deriving algorithms that can efficiently produce high-quality solutions (even if without approximation bounds) and/or optimality guarantees (even when the worst-case running time is exponential). In this work, we focus on the latter class of approaches, in particular the branch-and-bound procedure for the exact solution of Mixed Integer Programming (MIP) problems. MIP solvers are celebrated for their ability to solve problems with hundreds of thousands to millions of variables and constraints, a feat which the theory would suggest to be impossible in a reasonable amount of time. Yet, little is known about \textit{why} realistic instances of NP-Hard MIPs can be solved relatively efficiently in practice.

The existence of \textit{backdoor} sets provides one possible answer to this puzzle of empirical tractability in exact MIP solving. This was first reported by \citet{dilkina2009backdoors} who showed that backdoors of sizes 10--20 did exist for some instances (of MIPLIB2003~\citep{achterberg2006miplib}) with many hundreds to thousands of integer variables. The intuition is simple: if one need only branch on a handful of integer variables to solve a MIP instance to global optimality (or declare it as infeasible), then it is not difficult to imagine that a full-fledged MIP solver would be able to solve that instance without much branching. This motivates the design of algorithms for finding backdoors, with the following potential use cases:
    (i) Discovering heretofore unknown structural properties for a family of MIP instances, e.g., a new variant of an independent set or facility location problem, by data-mining backdoors from a large number of instances; 
    (ii) Constructing datasets of pre-computed backdoors that can then be used to train machine learning models for quickly identifying these crucial sets on similar but unseen instances and branching on them for a quick solving time;
    (iii) Directly using the backdoor for branching, assuming that the backdoor in question can be computed quickly in advance; such a use case, if realized, could speed up solving times substantially for instances that are challenging for current branching strategies in state-of-the-art MIP solvers.

How, then, does one find backdoors? We will denote the desired backdoor size by $K\in\mathbb{N}^{+}$. Despite negative computational complexity results (e.g.,~\citet{szeider2005backdoor} shows that finding backdoors for SAT requires time exponential in $K$, even for fixed $K$), some heuristics have been developed.  \citet{dilkina2009backdoors} used two forms of \textit{random sampling}: independent \textit{uniform} sampling of $K$ integer variables, and independent \textit{biased} sampling of $K$ integer variables to favor ones that are more fractional (i.e., with fractional parts closer to 0.5) in the solution of the Linear Programming (LP) relaxation of the MIP. Candidate sets are sampled (in parallel) and used for branching in a MIP solver; if global optimality is attained, then a backdoor has been found. The biased sampling strategy proved much more effective than uniform. While useful as a proof-of-concept, random sampling is a \textit{pure exploration} strategy which cannot use knowledge from previous draws to focus future ones towards more promising candidate sets.

\citet{fischetti2011backdoor} propose another strategy that leverages the following basic fact about (bounded) mixed 0--1 problems: the set of extreme points (vertices) of the LP-feasible region includes all integer-feasible points. The branch-and-bound tree that results from branching on a strong backdoor must have leaf nodes whose LPs are either infeasible or have integer-feasible optima. This leads to the following equivalent definition for a strong backdoor: a small set $\mathcal{B}$ of 0--1 variables such that each fractional vertex has at least one of its fractional variables in $\mathcal{B}$. 
If one could enumerate all (exponentially many) fractional vertices, then a Set Covering Problem (SCP) can be formulated to find a backdoor. In practice, the fractional vertices are collected progressively in batches by executing branch-and-bound using the current candidate backdoor. An SCP is solved at each iteration until all fractional vertices have been ``covered" or an early termination criterion is met. This method can be extended heuristically for general MIPs.  \citet{fischetti2011backdoor} demonstrate that this method's candidate backdoors, which cover many but rarely all fractional vertices, can lead to smaller trees if selected as branching variables early on in the search. However, because (i) there can be many fractional vertices and (ii) the order in which they are explored is arbitrary (see~\citep{fischetti2014exploiting, fischetti2014isco} for a discussion), it is unclear if the information that this method uses is actually conducive to backdoors in practice, i.e., it may be \textit{``exploiting" too aggressively}. Additionally, if one is interested in backdoors of size at most $K$ (as is the case in this paper), this approach may be unable to produce a solution as the SCP requires covering all of the collected fractional vertices.

\textbf{Contributions} Thus far, we have argued  (i) that finding backdoors is beneficial for a variety of tasks and (ii) that existing algorithms either explore or exploit too much. Towards a more general, effective, and extensible backdoor search algorithm, we propose \bamcts~(short for ``Backdoor MCTS"), a Monte Carlo Tree Search (MCTS) framework for finding backdoors to MIPs. Our contributions can be summarized as follows:

\begin{enumerate}
    \item \textbf{Backdoor Search as MCTS:} We contribute the first such formulation of the problem. \bamcts~can balance exploration and exploitation by design, is conceptually simple and easy to implement, and is extensible in a plug-and-play fashion.
    
    \item \textbf{Tight Integration with MIP Domain Knowledge:} To enable a scalable and effective solution, we customize the high-level MCTS procedure to the MIP setting through the use of domain-specific reward functions, action scoring functions and elimination rules, and careful engineering that is tightly coupled with CPLEX, a widely used MIP solver.
    
    \item \textbf{Extensive Empirical Evaluation:} reveals that \bamcts~ vastly outperforms the sampling strategy of~\citet{dilkina2009backdoors}, finding ``better" backdoors in a shorter amount of time on instances from the MIPLIB2017 Benchmark set~\citep{gleixner2021miplib}. We also show that branching on such sets of variables results in smaller search trees or optimality gaps compared to CPLEX with its default branching. 
\end{enumerate}

\section{Related Work}

\paragraph{Backdoors for Combinatorial Problems}
The notion of backdoors was first introduced by \citet{williams2003backdoors} for SAT, 
where it was observed that practical SAT instances often have a small tractable structures. Over the years many approaches
were proposed to find backdoors in the SAT context \citep{paris2006computing, kottler2008computation, li2011finding}. Observing the connection between SAT
and MIP, \citet{dilkina2009backdoors} generalized the concept of backdoors from SAT to MIP and proposed random sampling as a method for finding backdoors. \citet{fischetti2011backdoor} proposed another strategy for MIP, which to our knowledge remains the state-of-the-art to this day in the MIP context.

\paragraph{MCTS and Applications}
MCTS has seen a surge of interest thanks to its great success particularly in solving two-player games \citep{silver2017mastering}. 
This has led to many attempts in solving combinatorial optimization problems using MCTS by translating the problem into a game. In particular, UCT
has been used to guide MIP solvers \citep{sabharwal2012guiding} and to solve Quantified Constraint Satisfaction Problems (QCSP) \citep{satomi2011real}. \citet{bertsimas2014comparison} applied MCTS to the 
challenging problem of Dynamic Resource Allocation (DRA) and showed that it can greatly improve problem-specific baselines
on large scale instances. The Travelling Salesman Problem (TSP), another classical problem, was addressed in \citet{rimmel2011optimization} via a version of 
Monte Carlo search with some success. Bandit Search for Constraint Programming (BaSCoP) \citep{loth2013bandit} applied MCTS to improve the tree-search heuristics
of Constraint Programmin (CP) solvers and showed significant improvements on the depth-first search on certain CP benchmarks. More recent works involve successful application of the ``neural'' MCTS of AlphaZero to solve a variety of NP-Hard graph problems \citep{abe2019solving}, as well as solving First-Order Logic descriptions of combinatorial problems \citep{xu2021first}. 

The difference between our setting and a typical combinatorial optimization problem is that our reward function is an expensive black box, namely one that requires running a MIP solver for a limited number of steps, as opposed
to an analytical objective function; this makes our setting more challenging because evaluations are time-consuming.

\section{Technical Background}
\subsection{Branch-and-Bound for MIP}
We are concerned with Mixed Integer Linear Programming (MIP) problems of the form:
\begin{equation*}\label{eq:mip} z^* = \min\{c^Tx|Ax\leqslant b, x\in\mathbb{R}^n, x_j\in\mathbb{Z}\; \forall j\in I\}, \end{equation*}
where $A\in\mathbb{R}^{m\times n}$, $b\in\mathbb{R}^m$, $c\in\mathbb{R}^n$, and the non-empty set $I\subseteq \{1,...,n\}$ indexes the integer variables.
The vectors in the set $X_{MIP} = \{x\in\mathbb{R}^n|Ax\leqslant b, x_j\in\mathbb{Z}\; \forall j\in I\}$ are \emph{integer-feasible solutions}. An integer-feasible (or simply feasible) solution $x^*\in X_{MIP}$ is \emph{optimal} if $c^Tx^* = z^*$.

A MIP can be solved by \textit{branch and bound}~\citep{cook2012markowitz}, an exact algorithm that divides the original MIP into subproblems organized in a binary tree; see~\citet{WolseyNemhauser88} for a textbook exposition. At each node of the tree, an LP relaxation of the sub-problem is solved. If the resulting solution $x_N$ of the LP relaxation at a node $N$ is integral, then it is also a feasible solution to the MIP, i.e., $x_N\in X_{MIP}$. If such an integral solution has an objective value that is better than the best one found so far, it is referred to as the \textit{incumbent}, maintaining that designation until a better solution is found. Otherwise, the node is either pruned, if its lower bound is greater than the incumbent's value, or branched on, resulting in two child nodes that are added to the queue of nodes to be processed.

Pseudocosts are historical quantities, aggregated for each variable during the search, that represent the amount by which a node's LP relaxation value has been tightened when branching on a given variable. A higher pseudocost score indicates that branching on a variable typically helps make progress towards proving optimality. As such, pseudocosts are at the heart of most typical branching strategies that are used in MIP solvers~\citep{AchKocMar05,AchterbergBerthold09,Hendel15}. 

\subsection{Backdoors for MIP}
Given a MIP instance defined by the tuple $(A,b,c,I)$, a \textit{strong backdoor} of size $K\lll |I|$ is a set of integer variables $B, |B|=K, B\subset I$ such that branching exclusively on variables from $B$ results in a provably optimal solution or a proof of infeasibility. We will consider \textit{order-sensitive} strong backdoors, where $B$ is an ordered set: a variable at rank $i$ must be branched on before variables of rank $j>i$. 
The order in which the backdoor variables are considered for branching can affect the performance of the solver's primal heuristics, which in turn affects pruning in branch-and-bound. We refer to~\citet{dilkina2009backdoors} for a detailed discussion of why order matters in practice in MIP solvers. Throughout the paper, we will use the term ``candidate backdoor" to refer to an ordered set of integer variables that is being assessed but that may or may not be a strong backdoor.

\subsection{Monte Carlo Tree Search (MCTS)}
MCTS is a randomized algorithm for sequential games with a finite horizon and a finite number of actions $n$~\citep{browne2012survey}. At every step of a two-player game, MCTS seeks to identify the next action to take so as to maximize the probability of winning the game. It does so by building out an $n$-ary tree in which the root node represents the current state of the game, each edge represents a valid action, and a child node represents the extension of its parent's state by playing the action of the corresponding edge. Associated with each terminal state is a scalar reward value. 

Because the complete search tree is exponential in size, MCTS is organized into four key steps that can together navigate the exploration-exploitation trade-off in a sensible way, building out only a partial search tree that is sufficient for identifying a good next action. The four steps are:
\begin{enumerate}[(a)]
    \item\textbf{Selection:} 
    Given the current search tree, this step deals with \textit{selecting} the nodes in a depth-first dive from the root. Upper Confidence Trees (UCT)~\citep{KocsisSzepesvari06} is a widely used scoring rule that combines the average observed reward of a node (or state) with a function of the number of visits to the node; nodes with large average rewards and/or small visit counts obtain high UCT scores, balancing exploration and exploitation. The latter is controlled by an appropriately tuned hyperparameter.
    
    \item\textbf{Expansion:}
    When the selected node has only a subset of its potential children as nodes in the current search tree, one can choose to \textit{expand} the selected node's child set by creating a node for a new action. When the number of possible actions is large, Progressive Widening~\citep{coulom2007computing,couetoux2011continuous} is used to limit the branching factor of the tree and focus on expanding only frequently-visited (likely more promising) nodes. To select a new action to expand with, uniform sampling may be used. A more ``exploitative" expansion would select an action that has led to high-reward children in other nodes of the search tree. 
    
    \item\textbf{Simulation:}
    Following the depth-first dive from the root through a sequence of selections (and possibly a final expansion), a non-terminal node may be reached. Because rewards are only observed in terminal states, e.g., when the game concludes and a winner is declared, \textit{simulation} is used to traverse the state space from the node in question to a terminal state, without consolidating this subpath into the search tree. Typically, one picks actions randomly until a terminal state is reached.
    
    \item\textbf{Backpropagation:}
    Following the simulation, a reward is collected. \textit{Backpropagation} refers to the credit assignment process whereby the reward is passed on to the nodes along the depth-first path in the tree that led to the observed reward. A sum-backup rule is commonplace: each node accumulates rewards every time it is selected. Alternatively, a more aggressive max-backup rule keeps track of the maximum observed reward, see~\citep{sabharwal2012guiding} for example.
    
\end{enumerate}
The four-step loop is repeated until a termination condition (e.g., time limit) is reached, following which the action corresponding to the highest-reward or most visited child of the root node is ``played". The opponent responds, and the process is repeated starting with a new search tree representing the updated state of the game.

\section{\bamcts\footnote{Our implementation of \bamcts can be found at: \url{https://github.com/lyeskhalil/backdoorsearch}}}
Rather than treat MCTS as a black-box algorithm, we instantiate it for the backdoor search setting by incorporating as much domain knowledge about MIP solving as possible. 

We are given a MIP instance defined by the tuple $(A,b,c,I)$ and assume a user-defined bound on the backdoor size, $K\in\mathbb{N}^{+}$; $K\lll |I|$ should be seen as a small constant on the order of 5 to 10, which would imply a branch-and-bound tree with hundreds of nodes. 

We view backdoor search as a \textit{single-player}, \textit{deterministic} constraint satisfaction game. The goal is to find an order-sensitive strong backdoor that satisfies the size-$K$ constraint on the backdoor size.

We now define two key elements of MCTS for the backdoor setting. Let $P(I,K)$ denote the set of all permutations of all subset of the integer set $I$ that have size at most $K$. For example, if $K=2$ and $I=\{1,2,3\}$ then $P(I,K)=\{(),(1),(2),(3),(1,2),(1,3),(2,1),(2,3),(3,1),(3,2)\}$.
\begin{itemize}
    \item \textbf{State:} A state $S\in P(I,K)$ is a permutation of the variables s.t. $|S|\leq K$. A state is terminal iff $|S|=K$. The root of the MCTS tree corresponds to the empty state $()$.  
    
    \item \textbf{Action:} Given a non-terminal state $S$, a valid action is the index $i\in I$ of an integer variable such that $i\notin S$. Taking action $i$ in state $S$ means that $i$ is appended to the end of the ordered set $S$, leading to a new state $S'\in P(I,K), |S'|=|S|+1$ in which the first $|S|$ variables are the same as those in $S$.
\end{itemize}

\subsection{Building Blocks}

\paragraph{Candidate Evaluation} 
Consider a terminal state $\hat{S}$. To evaluate this candidate, a MIP solver is used to check if it is a backdoor. In particular, we instrument the CPLEX solver to (i) branch only on variables in $\hat{S}$, terminating the branch-and-bound if either the instance is solved (i.e., a backdoor has been found) or branching is no longer possible for at least one subproblem (a leaf node on the frontier of the branch-and-bound tree); (ii) respect the ordering implied by $\hat{S}$; (iii) collect auxiliary data such as pseudocosts and search completion information that will be useful for action selection and reward computation.

\paragraph{Reward Shaping} 
The goal -- finding an order-sensitive strong backdoor -- suggests a binary reward function in which a value of 1 is assigned to a terminal state during MCTS iff a backdoor is found, after which the search terminates because the goal has been achieved. However, this sparse reward structure makes any form of focused search impossible, an issue that we will tackle by using an appropriate reward function which assigns an informative score to a terminal state that is not a backdoor.

A good reward function is one that gives a large, but not maximal, value to a candidate which satisfies most of the conditions that define a strong backdoor. What are those ``conditions"? Following backdoor candidate evaluation using the MIP solver, a (potentially incomplete) branch-and-bound tree (not to be confused with the MCTS tree) can be observed. For the candidate to be a backdoor, all leaf nodes of the tree must be closed, i.e., they must have LP relaxations that are either infeasible, integer-feasible, or fathomed by bound; these are the conditions that must be simultaneously satisfied. Now consider a candidate which satisfies most but not all leaf node conditions. The \textit{tree weight}~\citep{kilby2006estimating} is a scoring function that maps a branch-and-bound tree to a value in $[0,1]$. When all integer variables are binary, the tree weight is defined as the fraction of binary assignments that belong to a closed leaf node; a strong backdoor would have a tree weight of 1 because all leaf nodes are closed and thus all binary assignments are covered. Importantly, the tree weight achieves our desired criterion for a good reward function: it can give meaningful scores to candidate backdoors that are not true strong backdoors but that eliminate many binary assignments. 

Consider the (binary) search tree of branch-and-bound for a mixed-binary integer program, and let $T_k$ denote the tree at the $k$-th iteration, i.e., after $k$ nodes have been expanded. Let $F_k$ denote the subset of tree nodes that are ``final" or fathomed due to their LP relaxations being infeasible, mixed-binary feasible, or worse in value than the best integer-feasible solution's value at the time they were expanded. The tree weight assigns to each node $v\in F_k$ a \textit{weight} of $2^{-d(v)}$, where $d(v)$ is the depth of $v$. The total tree weight of tree $T_k$ then writes:

$$\text{tree-weight}(T_k)=\sum_{v\in F_k}{2^{-d(v)}}.$$

At the start of branch-and-bound, no nodes have been fathomed and so $F_0=\emptyset, \text{tree-weight}(T_0)=0$. This function strictly increases with more fathomed nodes. We refer to section 4.3 of~\citep{hendel2020estimating} for further details and an illustrative example, and note that other tree search completion metrics therein could potentially be used instead of tree weight.  

\paragraph{The Role of Pseudocosts}
Pseudocosts are historical quantities aggregated for each variable during the search. The upwards (downwards) PC of a variable $x_j$ is the average unit objective gain taken over upwards (downwards) branchings on $x_j$ in previous nodes; we refer to this quantity as $\Psi_j^{+}$ ($\Psi_j^{-}$).
Pseudocost branching at node $N$ with LP solution $\check{x}$ consists in computing  values: 
$$	PC_j = \text{score} \Big((\check{x}_j - \floor{\check{x}_j}) \Psi_j^{-}, (\ceil{\check{x}_j} - \check{x}_j) \Psi_j^{+}\Big) $$
and choosing the variable with the largest such value. Typically, the product is used to combine the downwards and upwards values. One standard way to initialize the pseudocost values is by applying strong branching once for each integer variable, at the first node at which it is fractional~\citep{LinderothSavelsbergh99}. We will refer to this PC strategy with strong branching initialization as \emph{pseudocost branching} (PC).

As mentioned earlier, pseudocosts play a big role in most MIP branching strategies. Because the definition of a backdoor relies on branching, it is natural to consider ways in which pseudocosts can help steer \bamcts~ towards promising variables. Indeed, we will show a bit later how pseudocosts can serve as \textit{global scores} for actions (variables). Those global scores are then naturally incorporated into the selection and expansion steps of MCTS. For now, we emphasize that \bamcts~tracks the pseudocosts resulting from each backdoor candidate evaluation. We then maintain, for each variable, a running average of its pseudocost score across all candidate evaluations that involved the variable.

\paragraph{Action Space Reduction} 
We would like \bamcts~to scale to MIPs with tens or hundreds of thousands of integer variables. However, the MCTS tree grows fast with the number of integer variables, which may hamper progress. To reduce the action space, we leverage the empirical observation that only a few of the integer variables take on fractional values in the solution of the LP relaxation of the MIP; for instance,~\citet{berthold2014rens} shows that, on average across 159 instances from older MIPLIB instance libraries, 71.7\% of the integer variables are integer in the LP relaxation solution. Rather than work with the full integer set $I$, we restrict the action space to the subset $I_{\text{frac}}\subseteq I$ of variables that are fractional in the LP relaxation of the MIP instance. While heuristic, this restriction has some grounds in empirical MIP solving and reduces the action space dramatically. 

\subsection{Instantiating the Four Steps}

\paragraph{Selection}
We adopt a variant of the UCT selection rule, inspired by~\citet{gaudel2010feature}, that adds a global action score (average pseudocosts in our case) to UCT's typical elements. Consider an MCTS search tree node (or state) $S$ whose child node $S'$ is being assessed; assume $S'$ extends $S$ with variable $i\in I_{\text{frac}}$. 

We let $T_S$ denote the number of visits to node $S$; $\text{EXP}(S,S')$ denotes the exploration score of $S'$, which is large when $T_{S'}$ is much smaller than $T_S$; $\hat\mu_{S'}$ denotes the current average reward of state $S'$; $\mathbf{r}_{S'}$ is the vector of rewards that have been observed in the subtree rooted at $S'$ and $\sigma^2(\mathbf{r}_{S'})$ is its variance. 
Our final scoring function for node selection is given by~\eqref{eq:score}:
\begin{equation}
    \label{eq:score}
    \text{SCORE}(S, S')=(1-\alpha_{\text{PC}}) \text{UCT}_{\text{score}}(S, S') + \alpha_{\text{PC}} \hat{\text{PC}}_i.
\end{equation}

With $\alpha_{\text{PC}}\in[0,1)$, the scoring function is a convex combination of a UCT-type score for state $S'$ and the average pseudocost score $\hat{\text{PC}}_i$ of variable $i$. The latter may be interpreted as a RAVE score following~\citep{gelly2007combining}.
To arrive at the final scoring function, we define the exploration score (based on the standard UCT formula), the variance score (based on the UCB1-Tuned of~\citet{auer2002finite}), UCT without variance, UCT with variance, and the UCT score which is one of the two preceding scores depending on the value of \texttt{use\_variance}, respectively: 
\begin{equation*}
\begin{aligned}
\text{EXP}(S, S') &= \sqrt{\ln{(T_S)}/T_{S'}}\\
\text{VAR}(S') &= \sqrt{ \min\Big\{\dfrac{1}{4}, \sigma^2(\mathbf{r}_{S'}) + \text{EXP}(S, S')\Big\}}\\
\text{UCT}(S, S')&= \hat{\mu}_{S'} + C\cdot \text{EXP}(S, S') \\
\text{UCT}_{\text{var}}(S, S') &= \hat{\mu}_{S'} + C\cdot \text{EXP}(S, S') \cdot \text{VAR}(S')\\
\text{UCT}_{\text{score}}(S, S') &= 
\begin{cases}
      \text{UCT}_{\text{var}}(S, S') & \text{if \texttt{use\_variance}},\\
      \text{UCT}(S, S') & \text{otherwise}.
\end{cases}
\end{aligned}
\end{equation*}


To conclude, we note that the scoring function has three hyperparameters whose effects will be analyzed experimentally in the next section:

\begin{itemize}
    \item $\alpha_{\text{PC}}\in[0,1)$: the weight accorded to the global pseudocost average;
    \item $C\in\mathbb{R}_{>0}$: the exploration weight; 
    \item $\texttt{use\_variance}\in\{\text{True},\text{False}\}$: a boolean that determines whether UCT with or without variance is used.
\end{itemize}

\paragraph{Expansion}
Besides the traditional \textit{uniform random} expansion rule, we consider a deterministic \textit{best score} expansion rule which simply expands using the available action (variable) with the largest average pseudocost score. Together, these two rules cover a wide range along the exploration-exploitation spectrum.

\paragraph{Simulation}
We opt to proceed with random simulation as described in the preceding section.

\paragraph{Backpropagation}
We consider both the sum and max-backup rules here. While the latter is typically considered to be overly aggressive and wasteful (of reward information), it is quite suitable for our setting and allows for a form of focused local search: if a high-reward candidate has been observed in a node's subtree, a max-backup encourages more future visits to the same node. This may lead to slight modifications to the candidate that bring about improved rewards.

\subsection{Algorithm Engineering} 
Our implementation of \bamcts~exploits certain properties of the backdoor search problem to speed it up. One such property is that the root node of the branch-and-bound tree of each candidate evaluation is the same. Because solving the LP relaxation of the root node is typically much more time-consuming than other subproblems', we instrument CPLEX to solve the root LP once for all in advance and reuse its solution in subsequent evaluations. Rather than simulate and evaluate a single candidate at a time, we leverage the independence between the simulations to execute them and the candidate evaluation, in parallel.

\begin{figure*}[ht!]

  \centering
  \subfigure{\includegraphics[width=0.33\textwidth]{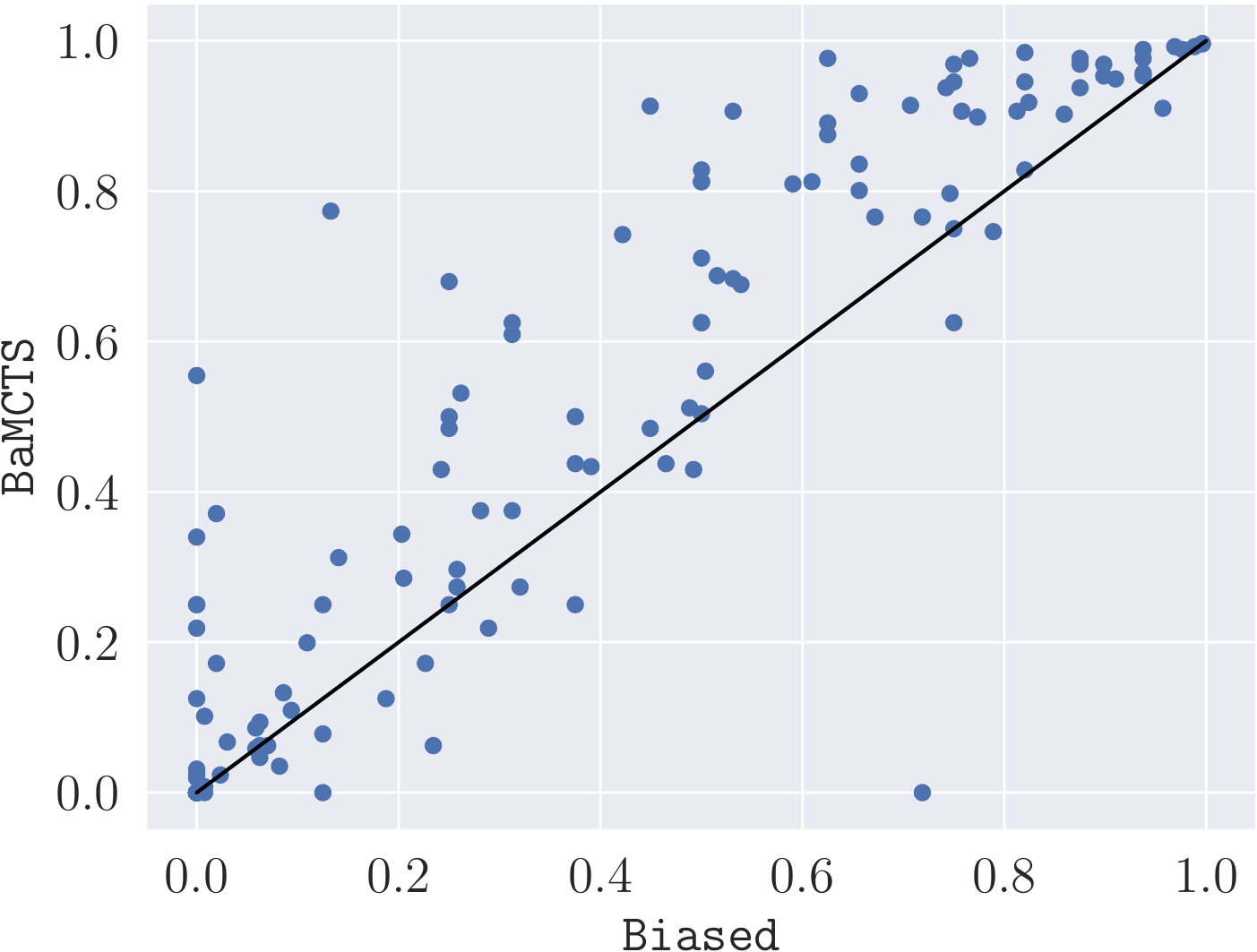}}
  \subfigure{\includegraphics[width=0.33\textwidth]{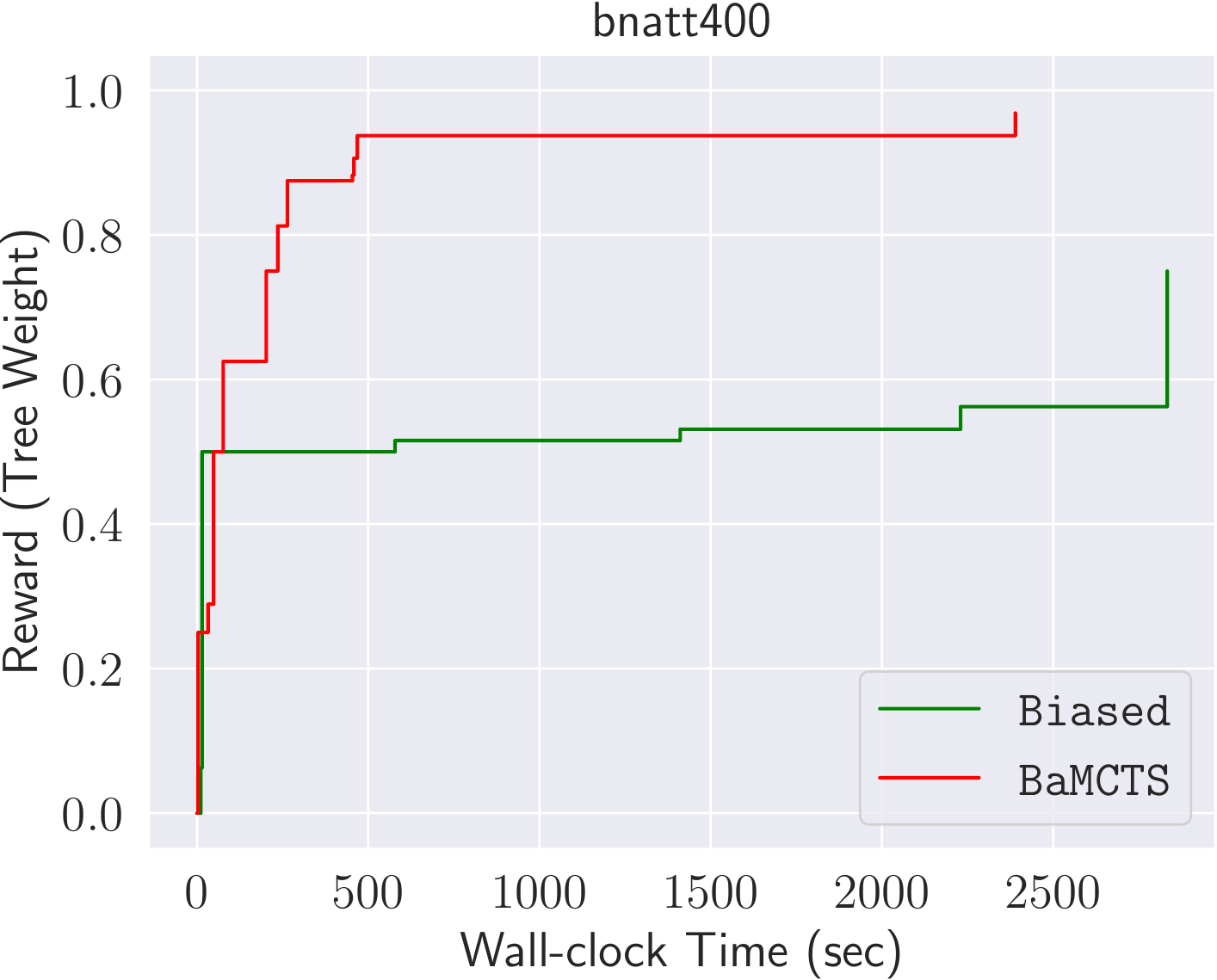}}
  \subfigure{\includegraphics[width=0.33\textwidth]{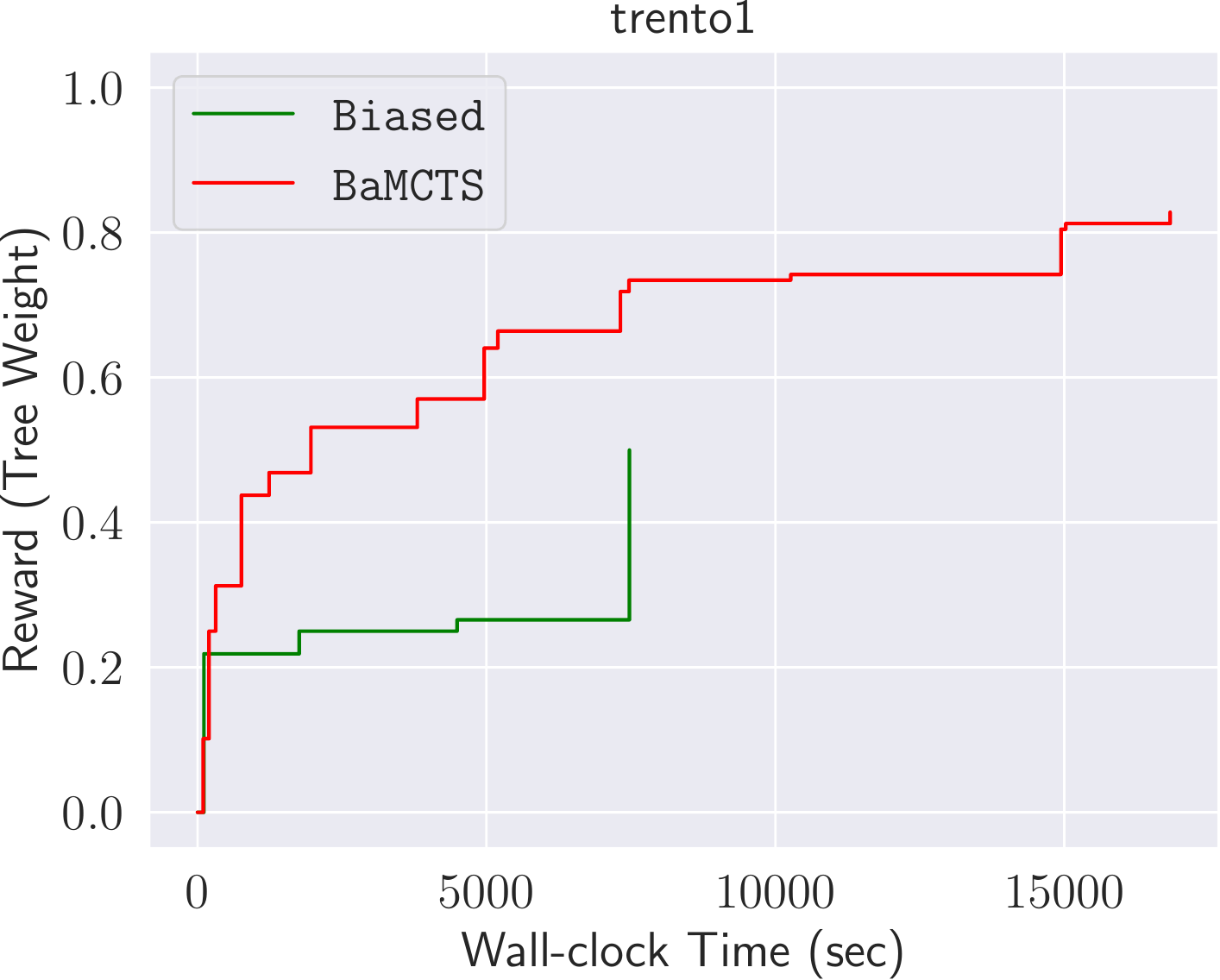}}
  
    \caption{ (left) \bamcts~outperforms \biased~in finding backdoors with higher (mean) reward: most instances fall above the diagonal. (middle) \& (right) \bamcts~attains higher rewards faster than \biased~on two representative instances.}
    \label{fig:mctsvssampling}

\end{figure*}

\section{Experiments}
To evaluate our method, we designed a set of experiments to answer the following questions:
\begin{enumerate*}[label=\textbf{\arabic*)}]
\item \textbf{Backdoor Extraction:} Can \bamcts~find backdoors with better tree weight values (or rewards) compared to the biased random sampling of~\citet{dilkina2009backdoors}?
\item \textbf{Sensitivity Analysis:} How sensitive is \bamcts~to its hyperparameters?
\item \textbf{Suitability of Tree Weight as a Reward Function:} Is branching on higher tree weight backdoors conducive to smaller search trees or (for instances that are not solved to optimality within a time limit) a smaller optimality gap? In other words, is tree weight a legitimate reward function for \bamcts?
\end{enumerate*}

\subsection{Experimental Setup}
\paragraph{Instances} We conducted our experiments on the MIPLIB2017 Benchmark set
\citep{gleixner2021miplib} that contains 240 instances. We only considered the 164 mixed-binary instances, i.e., instances with no general integer variables, due to the ease of implementation of the tree weight for binary problems; extension to general integer variables is possible. We presolved the instances in advance to eliminate redundant variables and constraints, and also let CPLEX generate cuts at the root node; the resulting preprocessed instance was used instead of the original in all subsequent procedure, as presolving and root cuts are both standard steps in MIP solving, indepedently of any backdoor search. Some instances were excluded due to either memory issues during presolving or being solved at the root node without branching, reducing the final dataset size down to 142 instances.

\paragraph{Baselines}
We compared \bamcts~against two baselines: \biased~and our implementation of \setcover~from \citet{fischetti2011backdoor}. Each run consisted of two phases: backdoor search and backdoor-guided MIP solving. 

\paragraph{Protocol for Backdoor Search} 
For the search phase, each method was given a budget of 5 hours per instance. We recorded the sequence of improving backdoors found by each method on every instance, particularly the backdoor with the highest tree weight. For a fair comparison, we tasked all methods to find backdoors of a specific size $K$. Naturally, it is trivial to find large backdoors (at the extreme, the full integer set is a backdoor) but their usefulness in the downstream MIP solving phase diminishes as they get larger. We report our results for backdoor size of $K=8$; however, we experimented with other backdoor sizes in the 5-10 range and the results still carry. 

\paragraph{Protocol for MIP Solving} In the second phase, we solved the same instance for one hour using CPLEX, but instrumented the solver to prioritize branching on the backdoor variables in the order they are given; this was achieved using CPLEX's branching priority feature.
To mitigate CPLEX's randomness w.r.t. arbitrary initial conditions~\citep{LodiTramontani13}, we solve each instance with 3 random seeds. 

\paragraph{\bamcts~Hyperparameters} To test the sensitivity of \bamcts, we sampled 30 hyperparameter configurations out of 128 configurations implied by the grid over:
\begin{itemize}
    \item Backup $\in \{\text{sum}, \text{max}\}$;
    \item Expansion Type $\in \{\text{Best Score}, \text{Uniform}\}$;
    \item $\texttt{use\_variance} \in \{\text{True}, \text{False}\}$;
    \item $\alpha_{\text{PC}} \in \{0.00, 0.01, 0.10, 0.50\}$;
    \item Exploration Parameter ($C$) $\in \{\frac{1}{\sqrt{2}}, 1, \sqrt{2}, 2, \sqrt{3}\}$.
\end{itemize}

For each configuration, we ran \bamcts~for one hour. We then selected the configuration with the highest mean tree weight. That single configuration (bottom row, Table~\ref{tab:configs}) was then used for the 5-hour backdoor search introduced earlier. 

\paragraph{Hardware} All experiments were conducted on a large CPU cluster. All MIP solving runs use a single core with an 8GB memory limit and a 1-hour time limit. Backdoor search runs for both \bamcts~and \biased~use 10 cores in parallel with a variable memory limit, capped at 63GB, that is proportional to the instance's size on disk, and a 5-hour time limit for the selected hyperparameter configuration. 

\subsection{Results}

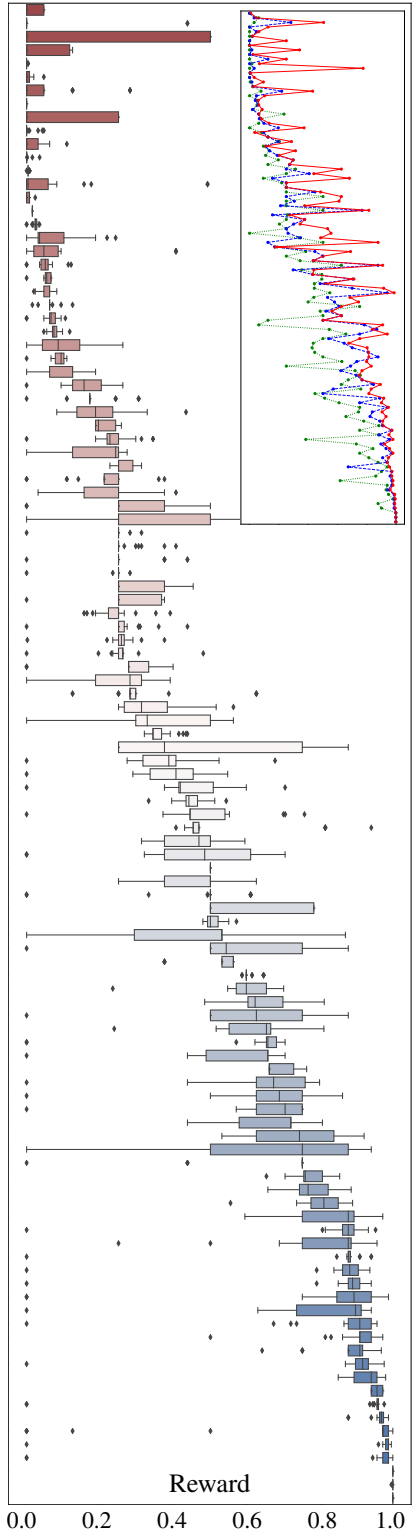
\begin{figure}[p]
    \centering
    \begin{tikzpicture}
    
    \node at (0, 0) {\includegraphics[width=.64\columnwidth]{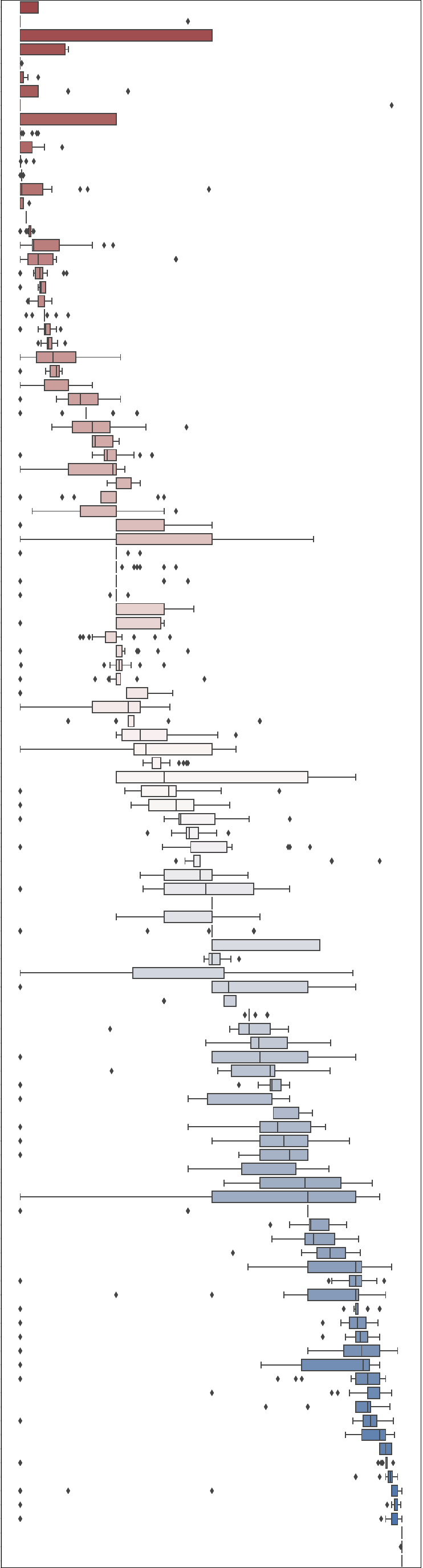}};
    \node at (1.5, 6.47) {\includegraphics[width=0.26\columnwidth]{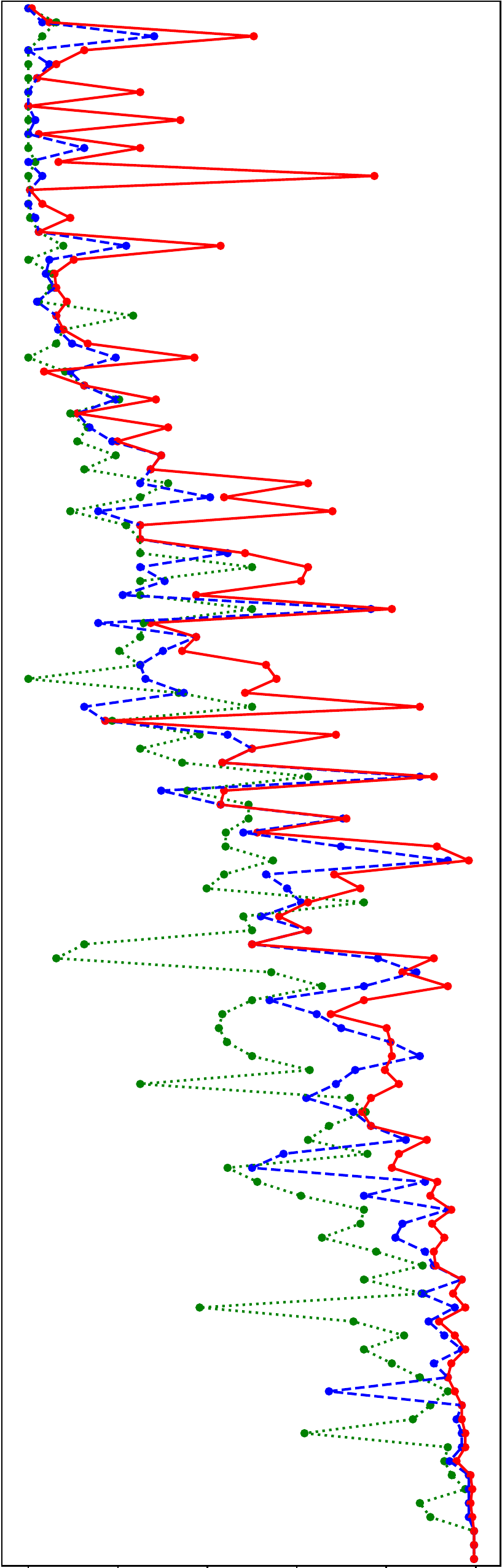}};
    
    \normalsize
    \draw (0, -9.7)  node {Reward};
    \small
    \draw (-2.5, -10.2)  node {0.0};
    \draw (-1.5, -10.2)  node {0.2};
    \draw (-0.5, -10.2)  node {0.4};
    \draw (0.5, -10.2)  node {0.6};
    \draw (1.5, -10.2)  node {0.8};
    \draw (2.4, -10.2)  node {1.0};
    \end{tikzpicture}
    \caption{Box-plots showing the distribution of rewards for each instance over the 30 hyperparameter configs. Instances are sorted based on median reward, hard (red) to easy (blue). \emph{top-right}: Reward comparison between \biased~(green) and \bamcts~run with 1 (blue) \& 5 (red) hours.}
    \label{fig:boxplot}
\end{figure}

\afterpage{
\noindent
\begin{minipage}{\columnwidth}
\centering

    \centering
    \begin{tikzpicture}
    
    \node at (0, 0) {\includegraphics[width=0.96\columnwidth]{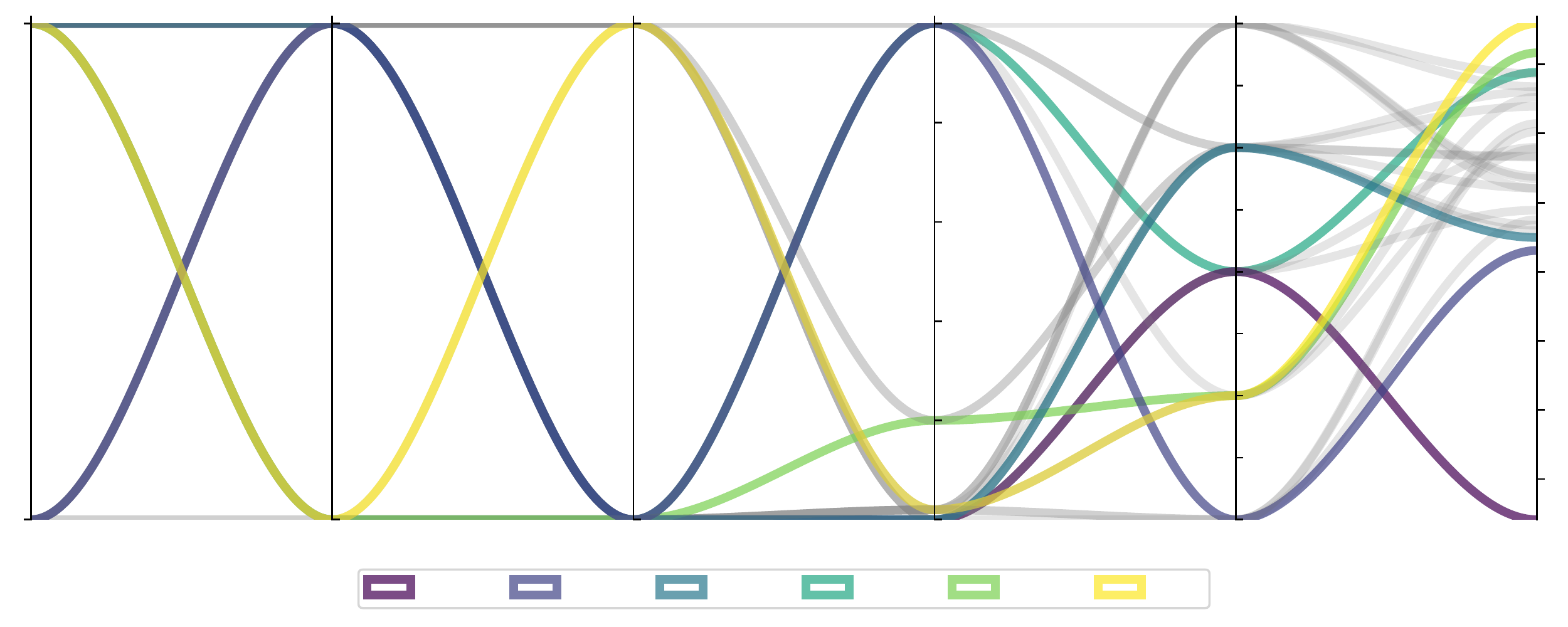}};
    
    \small
    \draw (-3.6, 2.1)  node {Backup};
    \draw (-2.2, 2.1)  node[align=center]{Expansion};
    \draw (-0.8, 2.05)  node[align=center]{\texttt{use\_}\\\texttt{variance}};
    \draw (0.8, 2.1)  node[align=center]{$\alpha_{\text{PC}}$};
    \draw (2.3, 2.1)  node[align=center]{Exp. $C$};
    \draw (3.6, 1.9)  node[align=center]    {Mean \\ Reward};
    
    \scriptsize
    \draw (-3.9, 1.65)  node {max};
    \draw (-3.9, -1.2)  node {sum};
    
    \draw (-2.3, 1.65)  node {uniform};
    \draw (-2.3, -1.2)  node {best score};

    \draw (-0.8, 1.65)  node {True};
    \draw (-0.8, -1.2)  node {False};
    
    \draw (3.64, -0.85)  node {0.22};
    \draw (3.64, 0.2)  node {0.28};
    \draw (3.64, 1.3)  node {0.34};

    \draw (2.5, -1.255)  node {$\frac{1}{\sqrt{2}}$};
    \draw (2.5, -0.43)  node {$1$};
    \draw (2.5, 0.3)  node {$\sqrt{2}$};
    \draw (2.5, 0.86)  node {$2$};
    \draw (2.6, 1.5)  node {$\sqrt{3}$};
    
    \draw (0.75, -1.2)  node {0.0};
    \draw (0.55, -0.55)  node {0.1};
    \draw (0.55, 0)  node {0.2};
    \draw (0.55, 0.53)  node {0.3};
    \draw (0.55, 1)  node {0.4};
    \draw (0.55, 1.5)  node {0.5};

    \draw (-1.75, -1.41)  node {10};
    \draw (-1, -1.41)  node {23};
    \draw (-.2, -1.41)  node {22};
    \draw (0.47, -1.41)  node {1};
    \draw (1.3, -1.41)  node {13};
    \draw (2.05, -1.41)  node {24};
    
    \end{tikzpicture}
    
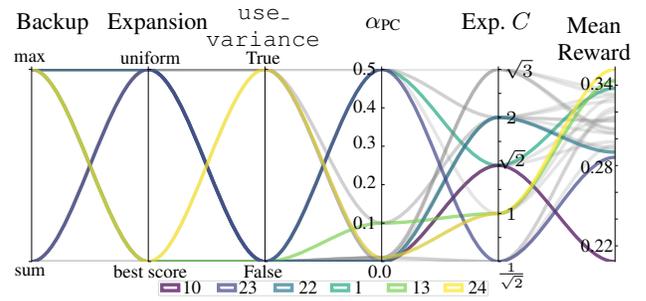
\captionof{figure}{\bamcts~is robust w.r.t. hyperparameter changes. Each curve in the figure represents a hyperparameter configuration (30 total) and indicates the values taken by that config as well as the obtained mean reward after running \bamcts~with that config on the dataset for 1 hour. Notice the concentration of rewards within the [0.28, 0.35] band. This shows that for most configs the reward does not change drastically. The highlighted curves represent the best and worst performing configs (see Table \ref{tab:configs} bellow).}
    \label{fig:besier}

\vspace{5mm}

\resizebox{\columnwidth}{!}{%
\begin{tabular}{@{}ccccccc@{}}
\toprule
  \multicolumn{1}{l}{} &
  \multicolumn{1}{l}{Backup} &
  \begin{tabular}[c]{@{}c@{}}Exp. \\ Type\end{tabular} &
  \begin{tabular}[c]{@{}c@{}}\texttt{use\_}\\\texttt{variance}\end{tabular} &
  $\alpha_{\text{PC}}$ & 
  $C$ &
  \begin{tabular}[c]{@{}c@{}}Mean\\ Reward\end{tabular} \\ \midrule
\textbf{10}    &   max &   Uniform &  False &    0.00 &    $\sqrt{2}$   &  0.208 \\
\textbf{23}    &   sum &   Uniform &  False &    0.50 &    $1/\sqrt{2}$ &  0.286 \\
\textbf{22}    &   max &   Uniform &  False &    0.00 &    $2$          &  0.289 \\
\textbf{20}    &   sum &   Uniform &  True  &    0.10 &    $2$          &  0.291 \\
\textbf{11}    &   max &   Uniform &  True  &    0.00 &    $2$          &  0.293 \\
\textbf{19}    &   max &   Uniform &  False &    0.01 &    $1/\sqrt{2}$ &  0.295 \\
\textbf{21}    &   sum &   Uniform &  False &    0.00 &    $\sqrt{2}$   &  0.297 \\
\textbf{16}    &   sum &   Uniform &  False &    0.50 &    $\sqrt{3}$   &  0.304 \\
\textbf{12}    &   max &   Uniform &  True  &    0.50 &    $2$          &  0.304 \\
\textbf{7}     &   max &   Uniform &  True  &    0.00 &    $\sqrt{3}$   &  0.306 \\
\textbf{3}     &   max &   Uniform &  False &    0.01 &    $\sqrt{3}$   &  0.307 \\
\textbf{9}     &   max &   Uniform &  True  &    0.10 &    $2$          &  0.313 \\
\textbf{0}     &   sum &   Uniform &  False &    0.01 &    $2$          &  0.313 \\
\textbf{17}    &   sum &   Uniform &  False &    0.50 &    $1$          &  0.315 \\
\textbf{6}     &   sum &   Uniform &  True  &    0.00 &    $\sqrt{2}$   &  0.315 \\
\textbf{26}    &   max &   Uniform &  True  &    0.01 &    $1/\sqrt{2}$ &  0.320 \\
\textbf{2}     &   max &   Uniform &  True  &    0.00 &    $1/\sqrt{2}$ &  0.322 \\
\textbf{4}     &   sum &   Best    &  False &    0.00 &    $2$          &  0.327 \\
\textbf{25}    &   max &   Best    &  False &    0.01 &    $1$          &  0.330 \\
\textbf{15}    &   max &   Best    &  True  &    0.50 &    $2$          &  0.332 \\
\textbf{5}     &   max &   Best    &  False &    0.01 &    $\sqrt{3}$   &  0.333 \\
\textbf{8}     &   sum &   Best    &  False &    0.01 &    $\sqrt{3}$   &  0.337 \\
\textbf{1}     &   max &   Best    &  False &    0.50 &    $\sqrt{2}$   &  0.337 \\
\textbf{13}    &   max &   Best    &  False &    0.10 &    $1$          &  0.343 \\
\textbf{24}    &   max &   Best    &  True  &    0.01 &    $1$          &  0.351 \\
\bottomrule
\end{tabular}}
\captionof{table}{ All \bamcts~parameter configurations sorted by mean reward (tree weight); higher is better.}
\label{tab:configs}
\vspace{5mm}
\end{minipage}
}%

\paragraph{Backdoor Extraction Performance} Figure \ref{fig:mctsvssampling} compares the performance of \bamcts~vs. \biased~in terms of the quality of the backdoors they find within the 5-hour time limit per instance. \bamcts~clearly outperformed \biased, leading to discovery of higher tree weight backdoors on many more instances. This is further demonstrated in the top-right plot of Figure \ref{fig:boxplot}. There, we observe that \bamcts~with even 1 hour time budget (blue) generally outperformed \biased~with 5 hour budget (green). \bamcts~with 5 hour budget (red) further widened the gap. Note that \setcover~does not depend on tree weight as a reward and is thus not comparable with \bamcts~here.

\paragraph{Sensitivity Analysis} The box-plots of Figure \ref{fig:boxplot} show the per-instance sensitivity of \bamcts~to the set of 30 hyperparameter configurations we tested. The instances are sorted by median reward from low (hard instances) to high (easy instances). Even though we observe quite high sensitivity for some instances, for the majority of them most configurations obtain similar reward. Figure \ref{fig:besier} shows all 30 hyperparameter configurations for \bamcts~as well as the average reward that each configuration achieved over the entire dataset. Most of the configurations resulted in average reward in the range of [0.28, 0.35], indicating that at least on average (across all instances) there is not a substantial variability in terms of the reward, and \bamcts~can be used out of the box with default hyperparameter values and achieve a reasonable performance. Note that MIPLIB2017 is quite diverse and it is likely that testing on a more homogeneous dataset would result in even lower parameter sensitivity. 



\begin{table}[t]

\resizebox{\columnwidth}{!}{%
\Huge
\begin{tabular}{@{}lcc|cc|cc@{}}
\toprule
& \multicolumn{2}{c}{seed 1 (47, 56, 4)} & \multicolumn{2}{c}{seed 2 (45, 61, 7)}  & \multicolumn{2}{c}{seed 3 (46, 60, 6)}  
\\ \midrule
& \STAB{\rotatebox[origin=c]{-60}{CPX--\texttt{def}}} & \STAB{\rotatebox[origin=c]{-60}{CPX--\bamcts}} & \STAB{\rotatebox[origin=c]{-60}{CPX--\texttt{def}}} & \STAB{\rotatebox[origin=c]{-60}{CPX--\bamcts}} & \STAB{\rotatebox[origin=c]{-60}{CPX--\texttt{def}}} & \STAB{\rotatebox[origin=c]{-60}{CPX--\bamcts}} 

\\ \midrule

\begin{tabular}[l]{@{}l@{}} \# of nodes \\ (solved by both)\end{tabular}             & 6902.7                                                & \textbf{6097.7}                         & 6173.8                                                & \textbf{5030.1}                         & 7744.9                                                & \textbf{7001.9}                         \\\midrule
\begin{tabular}[l]{@{}l@{}} total time \\ (solved by both) \end{tabular}                 & 179.5                                                 & \textbf{175.9}                          & 156.6                                                 & \textbf{138.3}                          & 169.4                                                 & \textbf{156.8}                          \\\midrule
\begin{tabular}[l]{@{}l@{}} optimality gap \\ (not solved by either) \end{tabular}           & 23/56                                                  & \textbf{33/56}                           & 24/61                                                  & \textbf{37/61}                           & 20/60                                                  & \textbf{40/60}                           \\\midrule
\begin{tabular}[l]{@{}l@{}}\# of instances \\  (solved by one method) \end{tabular} & \textbf{3}                            & 1                                                         & \textbf{4}                            & 3                                                         & 3                                                      & 3                                                         \\ \bottomrule
\end{tabular}%
}

\caption{Summary of MIP solving results. ``seed 1 (47, 56, 4)" means that for this seed, 47 instances were solved by both CPX--\texttt{def} and CPX--\bamcts, 56 were not solved by either within 1 hour, and 4 were solved by only one of the two. Geometric means are reported for ``\# of nodes" and ``total time"; the number of wins/losses are reported for ``optimality gap". The better method for each metric is in bold.}
\label{tab:mipresults}

\end{table}

\paragraph{Tree Weight and Branch-and-Bound Tree Size}
Table~\ref{tab:mipresults} summarizes the MIP solving results with a 1-hour time limit. CPX--\texttt{def} is CPLEX 12.10.0 with traditional branch-and-bound (i.e., with CPLEX's ``dynamic search" turned off); results with ``dynamic search" are similar. CPX--\bamcts~is CPLEX 12.10.0 with branching priorities defined by the best backdoor found for that instance by the best hyperparameter configuration of \bamcts~(configuration \textbf{24} in Table~\ref{tab:configs}). Of the initial 142 instances, we restrict the MIP solving here to 115 instances for which \bamcts~found at least one backdoor candidate with non-zero tree weight. Within each random seed, the instances are partitioned into three sets: (i) instances  solved by both methods: here we compare the shifted geometric means of the number of nodes and total time (lower is better; see page 33 of~\citep{Hendel15} for further discussion of the suitability of this metric), with shifts 100 and 10, respectively; (ii) instances that are not solved by either method: here the optimality gap after the 1-hour limit tells us which method made more progress;
we count the number of instances for which each method achieved a smaller gap (higher is better); 
and (iii) instances that are solved by only one of the methods: here we simply count the number of wins for each method (higher is better).

Table~\ref{tab:mipresults} shows that CPX--\bamcts~wins on the first three metrics consistently across the three seeds, with an average reduction of 700 to 1100 in the number of nodes. The last metric (final row) records two wins for CPX--\texttt{def} and one tie, but it is over 4 to 7 instances out of more than 110. We have also used the two-sided Wilcoxon signed-rank test, as described in~\citep{Hendel14}, to compare the distribution of values for the three metrics, and have found the corresponding p-values to be typically small, implying that the observed values for these metrics come from distributions with different medians for each of  CPX--\texttt{def} and CPX--\bamcts; this agrees with the conclusions from Table~\ref{tab:mipresults}.

\begin{figure}[t]
\centering
\includegraphics[scale=0.5]{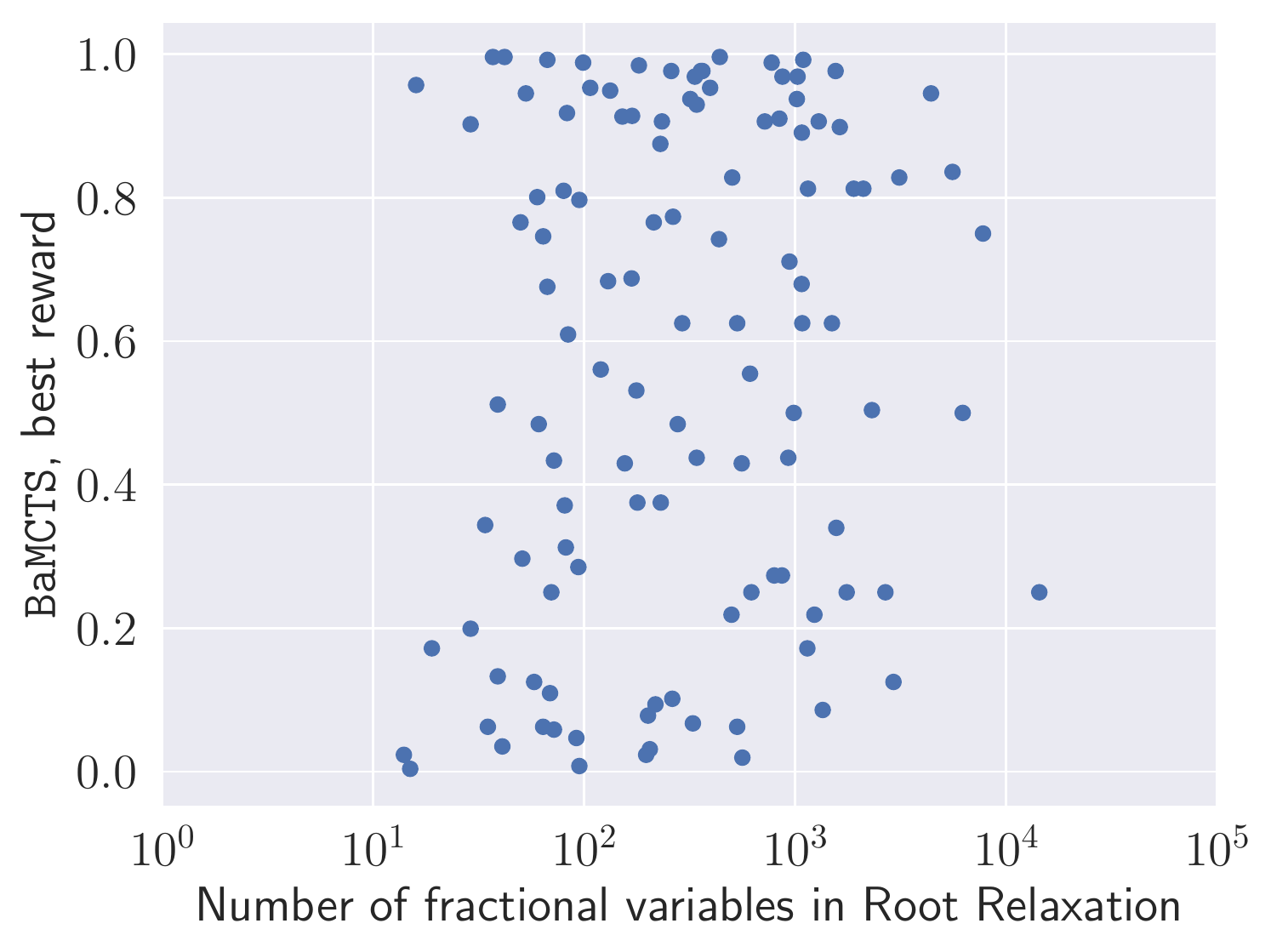}
\caption{Scatter plot of the number of fractional variables in the MIP root LP relaxation solution vs. the best reward (i.e., tree weight) found by our method.}
\label{fig:rewardvsnumfrac}
\end{figure}

The fact that branching with the best backdoor found by \bamcts~typically leads to better MIP solving (smaller number of nodes, smaller optimality gaps, etc.) confirms that tree weight is an informative reward signal for backdoor search.

Lastly, we note that we have attempted to compare against the \setcover~method of~\citet{fischetti2011backdoor} on the MIP solving metrics. However, \setcover~fails to return a size-8 backdoor for half of the 142 instances, requiring more variables to cover fractional vertices. This makes it difficult to compare \bamcts~to \setcover; modifications to the latter are necessary but are out of scope for this paper.

\paragraph{Impact on Total Time}
While Table~\ref{tab:mipresults} shows that branching with the backdoors found by our method result in only small reductions in total time, we note that: (1) this reduction is consistent across the three seeds and is in the 2-10\% range, which is non-trivial for standard MIPLIB2017 Benchmark instances; and (2) the experiment aims primarily at showing that branching with the backdoors translates into a smaller \textit{number of nodes}, a metric for which improvements are larger. Because CPLEX is not open-source, we are unable to control its behavior beyond its callbacks, but we believe that further total time speedups may be achieved with tighter integration with the internal solver code. 

\paragraph{Tree Weight vs. Number of Variables}
Given the wide range of tree weight values that were observed as a result of backdoor search, we were interested in whether only instances with small action spaces, i.e., a small number of integer variables that are fractional in the solution of the root LP relaxation of branch-and-bound tree, were associated with tree weights (rewards) that are close to 1. Figure~\ref{fig:rewardvsnumfrac} shows that is not the case at all: the tree weight values are spread across the wide range of values on the horizontal axis, hence our results are not biased towards ``easy" small instances.


\section{Conclusion}
In this paper we proposed \bamcts, a Monte Carlo Tree Search framework for finding strong backdoors in MIPs and demonstrated through our experiments its merits relative to earlier backdoor search algorithms in MIP literature. We hope that this paper would motivate further research in this direction and position MCTS as a viable approach in finding MIP backdoors. One particularly interesting future direction is to apply this technique on more homogeneous families of instances instead of MIPLIB to discover inherent structures for those problem families by studying their backdoors. We are planning to release our code to streamline these efforts.

\section*{Acknowledgements}
Elias B. Khalil acknowledges support from the Scale AI Research Chair Program and an NSERC Discovery Grant. Bistra Dilkina acknowledges support from NSF AI Institute for Advances in Optimization Award \#2112533.

\bibliography{references, referenceseliasaaai16}

\begin{thebibliography}{37}
\providecommand{\natexlab}[1]{#1}

\bibitem[{Abe et~al.(2019)Abe, Xu, Sato, and Sugiyama}]{abe2019solving}
Abe, K.; Xu, Z.; Sato, I.; and Sugiyama, M. 2019.
\newblock {Solving NP-Hard Problems on Graphs with Extended AlphaGo Zero}.
\newblock \emph{arXiv preprint arXiv:1905.11623}.

\bibitem[{Achterberg and Berthold(2009)}]{AchterbergBerthold09}
Achterberg, T.; and Berthold, T. 2009.
\newblock {Hybrid Branching}.
\newblock In \emph{CPAIOR}, 309--311. Springer.

\bibitem[{Achterberg, Koch, and Martin(2005)}]{AchKocMar05}
Achterberg, T.; Koch, T.; and Martin, A. 2005.
\newblock {Branching Rules Revisited}.
\newblock \emph{Operations Research Letters}, 33(1): 42--54.

\bibitem[{Achterberg, Koch, and Martin(2006)}]{achterberg2006miplib}
Achterberg, T.; Koch, T.; and Martin, A. 2006.
\newblock MIPLIB 2003.
\newblock \emph{Operations Research Letters}, 34(4): 361--372.

\bibitem[{Auer, Cesa-Bianchi, and Fischer(2002)}]{auer2002finite}
Auer, P.; Cesa-Bianchi, N.; and Fischer, P. 2002.
\newblock {Finite-Time Analysis of the Multiarmed Bandit Problem}.
\newblock \emph{Machine learning}, 47(2): 235--256.

\bibitem[{Berthold(2014)}]{berthold2014rens}
Berthold, T. 2014.
\newblock Rens.
\newblock \emph{Mathematical Programming Computation}, 6(1): 33--54.

\bibitem[{Bertsimas et~al.(2014)Bertsimas, Griffith, Gupta, Kochenderfer,
  Mi{\v{s}}i{\'c}, and Moss}]{bertsimas2014comparison}
Bertsimas, D.; Griffith, J.~D.; Gupta, V.; Kochenderfer, M.~J.;
  Mi{\v{s}}i{\'c}, V.~V.; and Moss, R. 2014.
\newblock {A Comparison of Monte Carlo Tree Search and Mathematical
  Optimization for Large Scale Dynamic Resource Allocation}.
\newblock \emph{arXiv preprint arXiv:1405.5498}.

\bibitem[{Browne et~al.(2012)Browne, Powley, Whitehouse, Lucas, Cowling,
  Rohlfshagen, Tavener, Perez, Samothrakis, and Colton}]{browne2012survey}
Browne, C.~B.; Powley, E.; Whitehouse, D.; Lucas, S.~M.; Cowling, P.~I.;
  Rohlfshagen, P.; Tavener, S.; Perez, D.; Samothrakis, S.; and Colton, S.
  2012.
\newblock A Survey of Monte Carlo Tree Search Methods.
\newblock \emph{IEEE Transactions on Computational Intelligence and AI in
  games}, 4(1): 1--43.

\bibitem[{Cook(2012)}]{cook2012markowitz}
Cook, W. 2012.
\newblock {Markowitz and Manne + Eastman + Land and Doig = Branch and Bound}.
\newblock \emph{Optimization Stories}, 227--238.

\bibitem[{Cou{\"e}toux et~al.(2011)Cou{\"e}toux, Hoock, Sokolovska, Teytaud,
  and Bonnard}]{couetoux2011continuous}
Cou{\"e}toux, A.; Hoock, J.-B.; Sokolovska, N.; Teytaud, O.; and Bonnard, N.
  2011.
\newblock Continuous Upper Confidence Trees.
\newblock In \emph{International Conference on Learning and Intelligent
  Optimization}, 433--445. Springer.

\bibitem[{Coulom(2007)}]{coulom2007computing}
Coulom, R. 2007.
\newblock Computing “Elo Ratings” of Move Patterns in the Game of Go.
\newblock \emph{ICGA journal}, 30(4): 198--208.

\bibitem[{Dilkina et~al.(2009)Dilkina, Gomes, Malitsky, Sabharwal, and
  Sellmann}]{dilkina2009backdoors}
Dilkina, B.; Gomes, C.~P.; Malitsky, Y.; Sabharwal, A.; and Sellmann, M. 2009.
\newblock Backdoors to Combinatorial Optimization: Feasibility and Optimality.
\newblock In \emph{International Conference on Integration of Constraint
  Programming, Artificial Intelligence, and Operations Research}, 56--70.
  Springer.

\bibitem[{Fischetti(2014)}]{fischetti2014isco}
Fischetti, M. 2014.
\newblock {BRANCHstorming (Brainstorming About Tree Search)}.
\newblock
  \url{http://www.dei.unipd.it/~fisch/papers/slides/2014%20ISCO%20%5bplenary%20Fischetti%20BRANCHstorming%5d.pdf}.
\newblock Accessed: 2022-04-13.

\bibitem[{Fischetti and Monaci(2011)}]{fischetti2011backdoor}
Fischetti, M.; and Monaci, M. 2011.
\newblock Backdoor Branching.
\newblock In \emph{International Conference on Integer Programming and
  Combinatorial Optimization}, 183--191. Springer.

\bibitem[{Fischetti and Monaci(2014)}]{fischetti2014exploiting}
Fischetti, M.; and Monaci, M. 2014.
\newblock Exploiting Erraticism in Search.
\newblock \emph{Operations Research}, 62(1): 114--122.

\bibitem[{Gaudel and Sebag(2010)}]{gaudel2010feature}
Gaudel, R.; and Sebag, M. 2010.
\newblock Feature Selection as a One-Player Game.
\newblock In \emph{International Conference on Machine Learning}, 359--366.

\bibitem[{Gelly and Silver(2007)}]{gelly2007combining}
Gelly, S.; and Silver, D. 2007.
\newblock Combining Online and Offline Knowledge in UCT.
\newblock In \emph{Proceedings of the 24th international conference on Machine
  learning}, 273--280.

\bibitem[{Gleixner et~al.(2021)Gleixner, Hendel, Gamrath, Achterberg, Bastubbe,
  Berthold, Christophel, Jarck, Koch, Linderoth et~al.}]{gleixner2021miplib}
Gleixner, A.; Hendel, G.; Gamrath, G.; Achterberg, T.; Bastubbe, M.; Berthold,
  T.; Christophel, P.; Jarck, K.; Koch, T.; Linderoth, J.; et~al. 2021.
\newblock MIPLIB 2017: Data-Driven Compilation of the 6th Mixed-Integer
  Programming Library.
\newblock \emph{Mathematical Programming Computation}, 1--48.

\bibitem[{Hendel(2014)}]{Hendel14}
Hendel, G. 2014.
\newblock \emph{Empirical Analysis of Solving Phases in Mixed Integer
  Programming}.
\newblock Master's thesis, Technische Universit{\"a}t Berlin.

\bibitem[{Hendel(2015)}]{Hendel15}
Hendel, G. 2015.
\newblock Enhancing MIP Branching Decisions by Using the Sample Variance of
  Pseudo Costs.
\newblock In \emph{Integration of AI and OR Techniques in Constraint
  Programming}, volume 9075, 199 -- 214.
\newblock In press.

\bibitem[{Hendel et~al.(2021)Hendel, Anderson, Le~Bodic, and
  Pfetsch}]{hendel2020estimating}
Hendel, G.; Anderson, D.; Le~Bodic, P.; and Pfetsch, M.~E. 2021.
\newblock Estimating the Size of Branch-and-Bound Trees.
\newblock \emph{INFORMS Journal on Computing}.

\bibitem[{Kilby et~al.(2006)Kilby, Slaney, Thi{\'e}baux, Walsh
  et~al.}]{kilby2006estimating}
Kilby, P.; Slaney, J.; Thi{\'e}baux, S.; Walsh, T.; et~al. 2006.
\newblock Estimating Search Tree Size.
\newblock In \emph{Proc. of the 21st National Conf. of Artificial Intelligence,
  AAAI, Menlo Park}.

\bibitem[{Kocsis and Szepesv{\'a}ri(2006)}]{KocsisSzepesvari06}
Kocsis, L.; and Szepesv{\'a}ri, C. 2006.
\newblock Bandit based Monte-Carlo Planning.
\newblock In \emph{ECML}, 282--293. Springer.

\bibitem[{Kottler, Kaufmann, and Sinz(2008)}]{kottler2008computation}
Kottler, S.; Kaufmann, M.; and Sinz, C. 2008.
\newblock Computation of Renameable Horn Backdoors.
\newblock In \emph{International Conference on Theory and Applications of
  Satisfiability Testing}, 154--160. Springer.

\bibitem[{Li and Van~Beek(2011)}]{li2011finding}
Li, Z.; and Van~Beek, P. 2011.
\newblock Finding Small Backdoors in SAT Instances.
\newblock In \emph{Canadian Conference on Artificial Intelligence}, 269--280.
  Springer.

\bibitem[{Linderoth and Savelsbergh(1999)}]{LinderothSavelsbergh99}
Linderoth, J.~T.; and Savelsbergh, M.~W. 1999.
\newblock A Computational Study of Search Strategies for Mixed Integer
  Programming.
\newblock \emph{INFORMS Journal on Computing}, 11(2): 173--187.

\bibitem[{Lodi and Tramontani(2013)}]{LodiTramontani13}
Lodi, A.; and Tramontani, A. 2013.
\newblock Performance Variability in Mixed-Integer Programming.
\newblock \emph{Tutorials in Operations Research: Theory Driven by Influential
  Applications}, 1--12.

\bibitem[{Loth et~al.(2013)Loth, Sebag, Hamadi, and
  Schoenauer}]{loth2013bandit}
Loth, M.; Sebag, M.; Hamadi, Y.; and Schoenauer, M. 2013.
\newblock Bandit-based Search for Constraint Programming.
\newblock In \emph{International Conference on Principles and Practice of
  Constraint Programming}, 464--480. Springer.

\bibitem[{Nemhauser and Wolsey(1988)}]{WolseyNemhauser88}
Nemhauser, G.~L.; and Wolsey, L.~A. 1988.
\newblock \emph{Integer and Combinatorial Optimization}.
\newblock John Wiley \& Sons.

\bibitem[{Paris et~al.(2006)Paris, Ostrowski, Siegel, and
  Sais}]{paris2006computing}
Paris, L.; Ostrowski, R.; Siegel, P.; and Sais, L. 2006.
\newblock Computing Horn Strong Backdoor Sets Thanks to Local Search.
\newblock In \emph{2006 18th IEEE International Conference on Tools with
  Artificial Intelligence (ICTAI'06)}, 139--143. IEEE.

\bibitem[{Rimmel, Teytaud, and Cazenave(2011)}]{rimmel2011optimization}
Rimmel, A.; Teytaud, F.; and Cazenave, T. 2011.
\newblock Optimization of the Nested Monte-Carlo Algorithm on the Traveling
  Salesman Problem with Time Windows.
\newblock In \emph{European Conference on the Applications of Evolutionary
  Computation}, 501--510. Springer.

\bibitem[{Sabharwal, Samulowitz, and Reddy(2012)}]{sabharwal2012guiding}
Sabharwal, A.; Samulowitz, H.; and Reddy, C. 2012.
\newblock Guiding Combinatorial Optimization with UCT.
\newblock In \emph{International conference on integration of artificial
  intelligence (AI) and operations research (OR) techniques in constraint
  programming}, 356--361. Springer.

\bibitem[{Satomi et~al.(2011)Satomi, Joe, Iwasaki, and Yokoo}]{satomi2011real}
Satomi, B.; Joe, Y.; Iwasaki, A.; and Yokoo, M. 2011.
\newblock Real-Time Solving of Quantified CSPs based on Monte-Carlo Game Tree
  Search.
\newblock In \emph{Twenty-Second International Joint Conference on Artificial
  Intelligence}.

\bibitem[{Silver et~al.(2017)Silver, Schrittwieser, Simonyan, Antonoglou,
  Huang, Guez, Hubert, Baker, Lai, Bolton et~al.}]{silver2017mastering}
Silver, D.; Schrittwieser, J.; Simonyan, K.; Antonoglou, I.; Huang, A.; Guez,
  A.; Hubert, T.; Baker, L.; Lai, M.; Bolton, A.; et~al. 2017.
\newblock Mastering the Game of Go Without Human Knowledge.
\newblock \emph{Nature}, 550(7676): 354--359.

\bibitem[{Szeider(2005)}]{szeider2005backdoor}
Szeider, S. 2005.
\newblock Backdoor Sets for DLL Subsolvers.
\newblock \emph{Journal of Automated Reasoning}, 35(1-3): 73--88.

\bibitem[{Williams, Gomes, and Selman(2003)}]{williams2003backdoors}
Williams, R.; Gomes, C.~P.; and Selman, B. 2003.
\newblock Backdoors to Typical Case Complexity.
\newblock In \emph{IJCAI}, volume~3, 1173--1178.

\bibitem[{Xu, Kadam, and Lieberherr(2021)}]{xu2021first}
Xu, R.; Kadam, P.; and Lieberherr, K. 2021.
\newblock First-Order Problem Solving through Neural MCTS based Reinforcement
  Learning.
\newblock \emph{arXiv preprint arXiv:2101.04167}.

\end{thebibliography}


\end{document}